\documentclass[11pt]{article}

\usepackage[utf8]{inputenc}
\usepackage[T1]{fontenc}
\usepackage[margin=1in]{geometry}
\usepackage{amsmath,amssymb}
\usepackage{booktabs}
\usepackage{caption}
\usepackage{xcolor}
\IfFileExists{lmodern.sty}
  {\usepackage{lmodern}\usepackage{microtype}}
  {\usepackage[expansion=false]{microtype}}
\usepackage[authoryear,round]{natbib}
\usepackage{hyperref}
\usepackage{xurl}
\usepackage{orcidlink}

\hypersetup{
  colorlinks=true,
  linkcolor=blue!55!black,
  citecolor=blue!55!black,
  urlcolor=blue!55!black
}
\newcommand{\DISCLOSURE}{The author led the development of the commercial appeal-processing product of which the described component is part.}

\title{\bfseries AI Appeals Processor: A Deep Learning Approach to Automated
Classification of Citizen Appeals in Government Services\thanks{Version~2
corrects the reference list (the operator accuracy of 67\% and the manual
handling times of 15--30 minutes are this study's own measurements),
corrects the parameter count of multilingual BERT, reports
per-domain F1 only, and adds confidence intervals, a description of the
production audit, and sections on lessons from deployment, limitations and
ethical considerations. All test-set results are unchanged.}}

\author{
Vladimir Beskorovainyi\,\orcidlink{0009-0004-7005-6242}\\
\small Besk Tech\\
\small \texttt{admin@besk.tech} \;\textbar\; \url{https://vladimir.besk.tech}
}

\date{}

\begin{document}
\maketitle
\begin{abstract}
Government agencies must register, classify and route every citizen appeal within statutory time limits, and much of this work is still done by hand. We describe AI Appeals Processor, a classification and routing component deployed in a CPU-only government environment, and report what its evaluation and deployment taught us. On 10,000 real Russian-language appeals from a cross-domain dataset, we compare Bag-of-Words and TF-IDF with SVM, fastText, Word2Vec+LSTM and multilingual BERT on a three-way appeal-type task. On a held-out test set of 1,500 appeals, BERT reaches 82\% accuracy and Word2Vec+LSTM 78\%, against 67\% for individual operators measured on an expert-adjudicated gold standard. We deployed the LSTM: in a workflow where an operator verifies every prediction, its lower training cost made frequent retraining on operator-verified labels practical, while the four-point accuracy gap did not change the operator's task. End-to-end handling time fell by 53--56\% across four appeal-length bands (unweighted mean 22.5 to 10.25 minutes); model inference takes under two seconds of this. Most residual errors trace to the label taxonomy rather than the model: the statutory definitions of complaints and applications overlap, and many appeals carry two intents. A post-deployment audit of production classifications, made after several retraining cycles by operators who saw the assigned category, judged more than 95\% correct; we explain why this figure is not comparable with the test-set result.
\end{abstract}


\section{Introduction}

Electronic channels have made it easy for citizens to write to public institutions, and every appeal must be registered, classified and routed to the responsible unit within time limits set by law. In the agency studied here, this was manual work. Measured against an expert-adjudicated gold standard, individual operators assigned the correct appeal type in 67\% of cases, and manual handling took 15 to 30 minutes per appeal depending on length.

A classifier that outperforms manual labelling on a held-out test set is only the first step. Deploying one is harder: the environment is a government data centre with a fixed budget and no GPU inference tier, and every decision must remain with a human officer. This paper reports a model comparison on 10,000 real appeals and, in more detail, the decisions and lessons that followed from deploying the second-best model into that environment.

Our contributions are: (1) a comparison of five classifiers on 10,000 real Russian-language citizen appeals under a fixed protocol, with a per-domain evaluation on seven thematic categories; (2) a description of a deployed CPU-only classification component in which operator review doubles as an annotation source for periodic retraining; (3) an error analysis showing that most residual errors stem from the label taxonomy, whose categories overlap by statutory definition; and (4) practical lessons on model selection, measurement and taxonomy design for public-sector text classification.

\section{Related Work}

\paragraph{Text classification in government.}
Supervised text classification can substantially reduce manual coding of political and administrative texts \citep{grimmer2013text}, and AI applications in the public sector face specific organisational and legal constraints \citep{wirtz2019artificial}. Recent public-sector NLP work has addressed related problems such as grouping large volumes of civic proposals by topic \citep{ferreira2025semantic}. Work on Russian-language citizen appeals under production constraints remains scarce.

\paragraph{Models and Russian NLP.}
We compare classical sparse models, fastText \citep{joulin-etal-2017-bag}, Word2Vec embeddings \citep{mikolov2013efficient} with an LSTM \citep{hochreiter1997long}, and BERT \citep{devlin-etal-2019-bert}; convolutional models are a common alternative for short texts \citep{kim-2014-convolutional}. Russian morphology is usually handled by lemmatisation with pymorphy2 \citep{korobov2015morphological}. Russian-specific encoders \citep{kuratov2019adaptation, zmitrovich-etal-2024-family}, sentence embedders evaluated on ruMTEB \citep{snegirev-etal-2025-russian}, and benchmarks such as RussianSuperGLUE and MERA \citep{shavrina-etal-2020-russiansuperglue, fenogenova-etal-2024-mera} now offer stronger starting points than multilingual BERT. Large language models can classify text with prompting \citep{sun-etal-2023-text}, although fine-tuned smaller models often remain competitive \citep{edwards-camacho-collados-2024-language}; we did not evaluate either.

\paragraph{Deployment and evaluation.}
Most of the effort in production ML lies outside the model \citep{sculley2015hidden, paleyes2022challenges}, and human-in-the-loop designs use human corrections as training signal \citep{wu2022survey}. On evaluation, we follow recommendations to report uncertainty and avoid over-reading small differences \citep{dror-etal-2018-hitchhikers, card-etal-2020-little}.

\section{System and Deployment Context}

\subsection{Overview}
The classifier is one component of a larger appeal-processing platform that also performs deduplication of related appeals arriving through different channels and retrieval of previous responses; this paper covers only classification and routing. The component is implemented as part of a microservice-based web application with five principal parts: (1) a user-facing frontend, (2) a backend server, (3) an AI module for text processing and classification written in Python with TensorFlow, (4) a relational database, and (5) an integration layer that connects to the agency's electronic document management system through a REST API. Components communicate asynchronously through a message broker. Classification results are written to the database and shown to an operator, who confirms or corrects the appeal type and destination before routing. No appeal is routed without human confirmation.

\subsection{Containerisation}
Each component runs in a container under an orchestrator with configurable replica counts. The production cluster is CPU-only: the AI module runs as three replicas with 2 CPU cores and 4~GiB of memory each, which provides availability and load distribution.

\subsection{Text Processing Pipeline}
\textbf{Preprocessing.} Text is lowercased and normalised, tokenised, stripped of NLTK Russian stop words and lemmatised with pymorphy2 \citep{korobov2015morphological}.
\textbf{Features.} Tokens are converted to TF-IDF vectors (5,000 features) or to 128-dimensional Word2Vec embeddings trained on the appeal corpus.
\textbf{Classification.} The production configuration uses the Word2Vec+LSTM model.
\textbf{Post-processing and feedback.} Results are passed to the backend for operator review and routing. Operator corrections are stored and periodically added to the training data for retraining.

\section{Data and Experimental Setup}
\label{sec:setup}

\paragraph{Data.}
We use 10,000 real citizen appeals from a fixed cross-domain dataset, spanning seven thematic domains, obtained under an institutional research agreement with a government agency. Personal data were removed before analysis, and the dataset cannot be redistributed. Each appeal has two labels: an \emph{appeal type} (complaint, application, proposal; 38\%, 36\% and 26\%) and a \emph{thematic domain} (seven categories, Table~\ref{tab:domain}). The three types are those defined in Article~4 of Federal Law No.~59-FZ \citep{fz59}. The legal definitions overlap: an \emph{application} covers a request for help in exercising one's rights but also a report of violations or deficiencies and criticism of public bodies, while a \emph{complaint} is a request to restore or protect violated rights. An appeal reporting a fault and asking for its repair can therefore satisfy both definitions.

\paragraph{Gold standard.}
Operators labelled appeals during routine processing; a panel of three domain experts then verified and corrected the labels (Fleiss' $\kappa=0.81$ among experts). The 67\% manual accuracy is that of individual operators' original labels on the test appeals against this gold standard.

\paragraph{Protocol.}
The data were split 70/15/15 (7,000/1,500/1,500), stratified by appeal type. All neural models were trained with the Adam optimiser and categorical cross-entropy loss. Configurations were selected by 5-fold cross-validation on the training set only; the selected configuration was retrained on the full training set, with the validation split used for early stopping. The test set was used once, for the final results in Tables~\ref{tab:main}, \ref{tab:confusion} and~\ref{tab:domain}. The models are:
\begin{enumerate}
\item \textbf{Operators}: labels assigned by individual operators during routine processing, evaluated against the gold standard.
\item \textbf{BoW+SVM}: term-count Bag-of-Words (5,000 features) with a linear SVM, $C$ selected from $\{0.1, 1, 10, 100\}$.
\item \textbf{TF-IDF+SVM}: TF-IDF features with a linear SVM and the same search space for $C$.
\item \textbf{fastText} \citep{joulin-etal-2017-bag}: supervised classifier with character 3--6-grams, learning rate 0.5, 25 epochs, embedding dimension 128, trained on non-lemmatised text after basic normalisation and stop-word removal.
\item \textbf{Word2Vec+LSTM}: 128-dimensional Word2Vec embeddings trained on the corpus, a single-layer unidirectional LSTM with 64 hidden units and dropout 0.3, a dense layer of 32 units (ReLU) and a softmax output; up to 50 epochs, batch size 64, learning rate 0.001, early stopping on validation loss with patience 5; maximum sequence length 256 tokens.
\item \textbf{mBERT}: fine-tuned \texttt{bert-base-multilingual-cased} \citep{devlin-etal-2019-bert} (about 178M parameters, most of them in the 119,547-token embedding matrix) with a classification head; 4 epochs, batch size 16, learning rate $2\times10^{-5}$ with linear warm-up, maximum sequence length 512 subword tokens.
\end{enumerate}
Experiments ran on a server with an Intel Xeon E5-2680 v4 CPU (2.40~GHz, 14 cores), 64~GiB RAM and an NVIDIA Tesla V100 GPU (16~GiB); training times in Table~\ref{tab:main} include model selection and refer to this server. We report accuracy and macro-averaged precision, recall and F1, with 95\% Wilson intervals for accuracy \citep{wilson1927probable}, computed from the rounded accuracies. Each neural result is a single training run.

\section{Results}

\begin{table}[htbp]
\centering
\begin{tabular}{lcccccc}
\toprule
Model & Accuracy & 95\% CI & Precision & Recall & F1 & Train (min) \\
\midrule
Operators (manual) & 67 & 64.6--69.3 & 67 & 65 & 66 & --- \\
BoW+SVM & 72 & 69.7--74.2 & 73 & 70 & 71 & 8 \\
TF-IDF+SVM & 75 & 72.7--77.1 & 76 & 73 & 74 & 12 \\
fastText & 76 & 73.8--78.1 & 77 & 75 & 76 & 3 \\
Word2Vec+LSTM & 78 & 75.8--80.0 & 78 & 78 & 78 & 95 \\
mBERT & 82 & 80.0--83.9 & 83 & 80 & 81 & 240 \\
\bottomrule
\end{tabular}
\caption{Appeal-type classification on the test set ($n=1{,}500$), \%. P, R and F1 are macro-averaged. Train: minutes, including model selection, on the experimental server.}
\label{tab:main}
\end{table}

\paragraph{Model comparison.}
Table~\ref{tab:main} shows that every automated model is more accurate than individual operators; for each, the Wilson interval lies above the operators' interval. Among the models, differences of one to three points (TF-IDF, fastText, LSTM) are within sampling uncertainty. BERT is four points above the LSTM, but the two intervals meet at 80.0\%, and we did not run a paired test such as McNemar's \citep{dietterich1998approximate}, so the size of this gap is uncertain. fastText reaches 76\% with three minutes of training and no lemmatisation, which suggests that character n-grams capture much of the morphological signal that lemmatisation supplies to the sparse models.

\begin{table}[htbp]
\centering
\begin{tabular}{lccc}
\toprule
True $\backslash$ Predicted & Complaint & Application & Proposal \\
\midrule
Complaint (570) & \textbf{445} & 86 & 39 \\
Application (540) & 72 & \textbf{410} & 58 \\
Proposal (390) & 33 & 42 & \textbf{315} \\
\bottomrule
\end{tabular}
\caption{Confusion matrix of Word2Vec+LSTM on the test set.}
\label{tab:confusion}
\end{table}

\paragraph{Where the errors are.}
Per-class recall is 78.1\% for complaints, 75.9\% for applications and 80.8\% for proposals (Table~\ref{tab:confusion}). Of the 330 errors, 158 (48\%) are confusions between complaints and applications, the pair whose statutory definitions overlap; 100 involve applications and proposals and 72 complaints and proposals. We manually inspected 100 randomly sampled misclassified appeals and assigned each to one dominant pattern. \emph{Dual intent} (42\%): the text contains two intents of different types, for example a report of a broken lift together with a request to repair it. \emph{Formal register} (31\%): standardised bureaucratic phrasing with no explicit marker of intent. \emph{Mixed framing} (27\%): a single intent expressed as a factual account with evaluative or emotional language, which resembles one type in form and another in content. The first pattern cannot be resolved by any single-label classifier; it calls for multi-label output or a separate intent head.

\begin{table}[htbp]
\centering
\begin{tabular}{lccc}
\toprule
Configuration & Accuracy & F1 & Train (min) \\
\midrule
\multicolumn{4}{l}{\textit{Embedding dimensionality (hidden = 64, dropout = 0.3)}} \\
\quad 64 & 75\% & 75\% & 18 \\
\quad 128 (selected) & \textbf{78\%} & \textbf{78\%} & 24 \\
\quad 256 & 77\% & 77\% & 38 \\
\midrule
\multicolumn{4}{l}{\textit{LSTM hidden units (dimension = 128, dropout = 0.3)}} \\
\quad 32 & 76\% & 76\% & 16 \\
\quad 64 (selected) & \textbf{78\%} & \textbf{78\%} & 24 \\
\quad 128 & 78\% & 78\% & 41 \\
\midrule
\multicolumn{4}{l}{\textit{Dropout rate (dimension = 128, hidden = 64)}} \\
\quad 0.1 & 76\% & 76\% & 22 \\
\quad 0.3 (selected) & \textbf{78\%} & \textbf{78\%} & 24 \\
\quad 0.5 & 77\% & 76\% & 26 \\
\bottomrule
\end{tabular}
\caption{Sensitivity of Word2Vec+LSTM, varying one factor at a time around the selected configuration. Training times are for single runs.}
\label{tab:sens}
\end{table}

\paragraph{Sensitivity.}
Table~\ref{tab:sens} varies one hyperparameter at a time around the selected configuration. Most differences are within sampling uncertainty. 256-dimensional embeddings do not help, plausibly because the corpus is too small to train them; 128 hidden units match 64 at 1.7 times the training time (41 vs.\ 24 minutes).

\begin{table}[htbp]
\centering
\begin{tabular}{lc}
\toprule
Domain & F1 \\
\midrule
Housing \& utilities & 84\% \\
Healthcare & 80\% \\
Education & 79\% \\
Transportation & 77\% \\
Urban development & 76\% \\
Social welfare & 74\% \\
Miscellaneous & 71\% \\
\midrule
\textit{Macro average} & \textit{77\%} \\
\bottomrule
\end{tabular}
\caption{Per-domain F1 of Word2Vec+LSTM on the seven-class domain task (test set), \%. Accuracy is 77\%.}
\label{tab:domain}
\end{table}

\paragraph{Thematic domains.}
On the separate seven-class domain task the LSTM reaches 77\% accuracy and 77\% macro F1 (Table~\ref{tab:domain}). Domains with specialised vocabulary score highest; the heterogeneous ``miscellaneous'' class scores lowest.

\begin{table}[htbp]
\centering
\begin{tabular}{lccc}
\toprule
Length (words) & Manual & Assisted & Reduction \\
\midrule
$<100$ & 15.0 & 7.0 & 53\% \\
100--300 & 20.0 & 9.0 & 55\% \\
300--500 & 25.0 & 11.0 & 56\% \\
$>500$ & 30.0 & 14.0 & 53\% \\
\midrule
Unweighted mean & 22.5 & 10.25 & 54\% \\
\bottomrule
\end{tabular}
\caption{Average end-to-end handling time per appeal (minutes), including operator review and routing.}
\label{tab:time}
\end{table}

\paragraph{Handling time.}
Table~\ref{tab:time} reports average end-to-end handling time per appeal, measured during the study period for the manual workflow and for the assisted workflow in which the operator reviews the model's suggestion. The reduction is 53--56\% in every length band. The bottom row is the unweighted mean of the four bands, not a mean weighted by the share of appeals in each band. Preprocessing and inference take under two seconds per appeal; nearly all of the remaining time is human verification.

\begin{table}[htbp]
\centering
\begin{tabular}{lc}
\toprule
Concurrent requests & Average handling time (min) \\
\midrule
1 & 10.25 \\
5 & 10.50 \\
10 & 11.20 \\
20 & 12.80 \\
50 & 15.30 \\
100 & 18.60 \\
\bottomrule
\end{tabular}
\caption{Average end-to-end handling time per appeal in the assisted workflow under concurrent load.}
\label{tab:load}
\end{table}

\paragraph{Behaviour under load.}
In a stress test of the assisted workflow (Table~\ref{tab:load}), average handling time per appeal rose by 25\% at 20 concurrent requests (from 10.25 to 12.80 minutes) and reached 18.60 minutes at 100. The increase comes from queueing at the operator review stage, which was staffed by a fixed pool of five reviewers, not from the model: inference stays under two seconds per appeal regardless of concurrency. Capacity planning for such a system is therefore a question of reviewer staffing more than of compute.

\section{Lessons from Deployment}
\label{sec:lessons}

\paragraph{We shipped the second-best model.}
BERT was the most accurate model, but we deployed the LSTM for three reasons. First, the production environment had no GPU inference tier. Second, training including model selection took 95 minutes for the LSTM against 240 for BERT on our experimental GPU server, and a single LSTM training run takes about 24 minutes, which made frequent retraining on new operator-verified labels practical. Third, because an operator reviews every prediction, a four-point accuracy difference did not change the operator's task: it changes how often a suggestion is corrected, not whether it is checked. We did not measure BERT's inference latency under the production limits. A Russian-specific compact encoder would be the first model we would now test.

\paragraph{Operator review is an annotation engine.}
Every confirmation or correction becomes a labelled example. Corrections feed periodic retraining, and the most useful engineering decision was to log the state of every field before and after the operator touched it. Without that log, corrections cannot be separated from routine confirmations, and retraining data cannot be audited.

\paragraph{The production audit and the test score measure different things.}
After several retraining cycles, a sample of production appeals was exported with the categories the system had assigned and checked item by item by operators, who judged more than 95\% of the classifications correct. This figure is not comparable with the 78\% test accuracy. It was measured on the live production stream rather than on the fixed set of 10,000 real appeals used for the model comparison, by operators who saw the assigned category rather than by blind relabelling, and on a model retrained on operator-verified labels. It is the production classification accuracy of the retrained model as judged by an operator audit.

\paragraph{The taxonomy is the bottleneck.}
About half of the errors fall between two legally overlapping categories, and dual-intent appeals have no correct single label. Further gains are more likely to come from multi-label output for intent and finer subcategories than from a larger model.

\paragraph{Measure the interface, not only the model.}
Inference is a negligible share of handling time, so the time saving comes from what the operator sees and how quickly it can be confirmed. We would now instrument the operator interface with per-appeal timestamps from the start, so that handling time can be analysed per appeal and not only as band averages.

\section{Conclusion}

On 10,000 real Russian-language appeals, all evaluated classifiers outperform individual operators; a Word2Vec+LSTM model reaches 78\% accuracy on the three-way appeal-type task and 77\% on seven thematic domains. We deployed this model in a CPU-only government environment because, with an operator verifying every prediction, cheap and frequent retraining on operator-verified labels mattered more than four points of test accuracy. Handling time fell by about half, almost all of the remaining time being human verification. Most residual errors come from the taxonomy itself. Future work will evaluate Russian-specific compact encoders and locally hosted language models under CPU-only limits, add multi-label intent, and report variance over repeated runs.

\section*{Limitations}

\emph{Data.} All results come from one proprietary Russian-language dataset that cannot be released. The split was random and stratified rather than temporal, and we do not report deduplication of repeated appeals, removal of web-form boilerplate or the treatment of attachments; near-duplicates across splits can inflate test scores \citep{lee-etal-2022-deduplicating}. Inputs were truncated to 256 tokens (LSTM) and 512 subword tokens (BERT).

\emph{Labels.} The gold standard was produced by experts correcting operators' labels, which may bias the estimate of operator accuracy upwards. Details of the agreement study behind $\kappa=0.81$ are not reported. The single-label taxonomy forces dual-intent appeals, and any requests for information or clarification, into one of three types.

\emph{Evaluation.} Each neural result is a single run without variance estimates, and no paired significance tests were performed. The NLTK Russian stop-word list removes negation particles and lemmatisation removes verbal mood, both potentially informative here; neither was ablated. Russian-specific encoders and language models were not evaluated, and BERT's inference cost under production limits was not measured.

\emph{Time and production figures.} The handling-time protocol, the number of appeals per band and per-band variance are not available, so Table~\ref{tab:time} is descriptive, and its summary row is an unweighted mean; the same applies to the load test in Table~\ref{tab:load}. The production audit was not blind, its sample size is not reported, and it is not comparable with the test-set results.

\section*{Ethical Considerations}

The data are citizens' appeals to a public authority and were used under an institutional agreement, with personal data removed before analysis; they are not released. The system is decision support: an operator confirms or corrects every prediction, and no appeal is routed without human confirmation. Misclassification can still delay an appeal, so the human check is a safeguard rather than a formality. We did not evaluate whether error rates differ across groups of citizens, regions or writing styles; such differences could translate into unequal delays and should be monitored in any deployment.

\section*{Disclosure}

\DISCLOSURE

\bibliographystyle{plainnat}
\bibliography{references}
\end{document}